\documentclass{article}

\usepackage[square]{natbib}
\usepackage{graphicx}
\usepackage{caption}
\usepackage{subcaption}
\usepackage{dcolumn}
\newcolumntype{d}[1]{D{.}{.}{#1}}
\usepackage{multirow}
\usepackage[papersize={8.5in,11in}, margin=1.2in]{geometry}

\setcitestyle{numbers}

\usepackage[utf8]{inputenc} 
\usepackage[T1]{fontenc}    
\usepackage{hyperref}       
\usepackage{url}            
\usepackage{booktabs}       
\usepackage{amsfonts}       
\usepackage{nicefrac}       
\usepackage{microtype}      

\title{Wide Boosting}

\author{%
  Michael T. Horrell, PhD \\
  \texttt{https://github.com/mthorrell}
}

\begin{document}

\maketitle

\begin{abstract}
Gradient Boosting (GB) is a popular methodology used to solve prediction problems by minimizing a differentiable loss function, $L$. GB performs very well on tabular machine learning (ML) problems; however, as a pure ML solver it lacks the ability to fit models with probabilistic but correlated multi-dimensional outputs, for example, multiple correlated Bernoulli outputs.  GB also does not form intermediate abstract data embeddings, one property of Deep Learning that gives greater flexibility and performance on other types of problems.  This paper presents a simple adjustment to GB motivated in part by artificial neural networks. Specifically, our adjustment inserts a matrix multiplication between the output of a GB model and the loss, $L$. This allows the output of a GB model to have increased dimension prior to being fed into the loss and is thus ``wider'' than standard GB implementations. We call our method Wide Boosting (WB) and show that WB outperforms GB on mult-dimesional output tasks and that the embeddings generated by WB contain are more useful in downstream prediction tasks than GB output predictions alone.
\end{abstract}

\section{Introduction}

Gradient Boosting (GB), first discussed in general in \cite{gbm}, is a popular methodology to build prediction models. Specifically, GB finds a function, $f$, restricted to a class of functions, $\mathcal{F}$, that attempts to minimize
\begin{equation}
L(Y,f(X)) \label{loss}
\end{equation}
for $L$ a differentiable loss function and $Y \in \mathbb{R}^{n \times d}$ and $X \in \mathbb{R}^{n \times p}$ and $f(x) \in \mathbb{R}^{n \times q}$. For the most common $L$, $q = d$. Gradient Boosting accomplishes this by composing $f$ out of additive, iterative adjustments, $a_i$. $f_0$ can be initialized to be a constant function. Then
\begin{equation}
    f_{i+1} = f_i + \eta_i a_i \label{iterupdate}
\end{equation} where $\eta_i$ is a scalar and $a_i$ approximates or is otherwise related to $-\frac{\partial L(Y,f_i(X))}{\partial f_i(X) }$; thus, GB can be viewed as a kind of gradient descent in function space.

\cite{xgboost} tabulated results from Kaggle, a prediction competition website, and found around 60\% of winning solutions posted in 2015 used a particular software package implementing GB, \texttt{XGBoost} \citep{xgboost}. In a review of the 2015 KDD Cup, \cite{kddreview} further remarks on the effective and popular use of \texttt{XGBoost} in that competition. Since 2015, other software packages relying on GB have been developed and open-sourced by researchers at Microsoft (\texttt{LightGBM} \citep{lightgbm}) and Yandex (\texttt{CatBoost} \citep{catboost}). \texttt{LightGBM} has also been effectively used in several Kaggle prediction competitions \cite{microsoft}.

In this paper, we develop a method that generalizes standard GB frameworks that we call Wide Boosting (WB). This method introduces a matrix, $\beta$, multiplied to $f(X)$ so that the output dimension of $f(X)$, $q$, can be arbitrary. The output of the usual GB model is therefore an intermediate embedding of the input data. Writing WB in terms of the overall loss, WB estimates $f$ and $\beta$ in attempt to minimize $L(Y, f(X)\beta)$.

We have implemented Wide Boosting in a Python package, \texttt{wideboost} (available via \url{pypi.org} or at \url{https://github.com/mthorrell/wideboost}). Notably, \texttt{wideboost} is able to use existing GB packages as a backend; thus we can take advantage of these highly optimized packages. Currently the \texttt{XGBoost} and \texttt{LightGBM} backends are implemented for \texttt{wideboost}. The \texttt{XGBoost} backend is currently the only backend supporting multi-dimesion responses (apart from multinomial response which is supported natively by \texttt{XGBoost} and \texttt{LightGBM}).

As a broader interpretation, Wide Boosting is exactly analogous to treating $f(X)$ as the output of the hidden layer in a dense, one-hidden-layer neural network. From this perspective, $f(X)$ embeds the rows of $X$ in a $q$-dimensional space prior to being processed for prediction. Other works combine the powerful neural network and tree fitting methodologies in different ways (\citep{dndf}, \citep{deepgbm}, \citep{ndt}, \citep{node}). However, these works adjust the base methodologies and introduce further complexities to merge the approaches. Wide Boosting leverages both boosting and neural network architecture methodologies without significant adjustment and hence is able to take direct advantage of improvements in world-leading implementations of boosting while gaining some properties of neural networks. As one new method combining two popular prediction paradigms, Wide Boosting potentially points to more research areas where nodes of computational networks have new structures and methods to fit them. As an additional example, \cite{blog} indicates that the weights in a feed-forward neural networks can be usefully fit using boosting. 

It should also be noted that the simple $\beta$ multiplication isn't the only way to merge GB and neural networks. Indeed GB only requires a differentiable function relating inputs to a response (preferably twice differentiable to use \texttt{XGBoost} and \texttt{LightGBM} as a backend); thus, for example, gradient boosted decision trees can be put as the first layer in any feed-forward neural network.

The rest of this paper is structured as follows. Section \ref{wbreview} details Wide Boosting, its parameters and necessary calculations. Section \ref{exp} reviews numerical experiments using \texttt{wideboost}.

\section{Wide boosting}
\label{wbreview}

Let $f(X) \in \mathbb{R}^{n \times q}$ and let $\beta \in \mathbb{R}^{q \times d}$. We fit $f$ and $\beta$ in attempt to minimize
\begin{equation}
L(Y,f(X) \beta). \label{wb}
\end{equation}
With this formulation, $q$ is the output dimension of $f(X)$, and $\beta$, using matrix multiplication, gives $f(X)\beta$ a dimension that matches $Y$. 

If we think of $\beta$ as part of the loss function, wide boosting is a form of gradient boosting and can make use of existing theory and methods to find $f$. Namely $f$ can be additive decision trees, the usual method of estimating $f$ in GB models.  If $L(Y,\cdot)$ is convex in its second argument, $L(Y,\cdot \beta)$ is convex also. This property further preserves some convergence properties of note in \cite{grubb2011generalized}. For major existing implementations of gradient boosting (\texttt{XGBoost} \citep{xgboost}, \texttt{LightGBM} \citep{lightgbm}, \texttt{CatBoost} \citep{catboost}), wide boosting can be implemented by simply providing these frameworks with the right gradient and hessian calculations. We give general calculations of gradients and hessians in Appendix \ref{gradhes}.

\subsection{Constructing and estimating $\beta$}
\label{constb}

We tried a few methods of initializing $\beta$ and have left these methods as hyperparameters in the \texttt{wideboost} package. Using an identity matrix appended to a uniform random matrix or a purely uniform random matrix make up our base $\beta$ initializations. A further adjustment can be made to normalize the columns of these matrices in order to decouple the ``wideness'' of the network from the usual step parameter, $\eta$. Note that when $\beta = I$ and $q=d$, WB is equivalent to GB.

Like $f$, $\beta$ can also be estimated. $\beta$ can be estimated in the same manner that one can use boosting to estimate linear regression coefficients. $\hat{\beta}$ is sequentially updated each round by fitting adjustment coefficients to $-\frac{\partial L(Y,f_i(X))}{\partial f_i(X) }$.  Many other methods could be used to estimate $\beta$, but fitting $\beta$ also via boosting is convenient because the same intermediate gradients get calculated when estimating $f(X)$ as well. As a note of caution, initial experiments showed fitting $\beta$ can be fairly unstable. This might be improvable through more careful conditioning. Currently in $\texttt{wideboost}$, using very small learning rate for $\beta$ best handles these stability issues. Moreover, not fitting $\beta$ (setting a learning rate to 0) does not seem to cause a large performance drop.

\subsection{Computational Scaling and Efficiency}

Given our main implementation of WB makes use of \texttt{XGBoost} or \texttt{LightGBM} software packages as computational backends, we inherit both the optimizations employed in those packages and their limitations. Specifically both \texttt{XGBoost} and \texttt{LightGBM} fit additional, independent trees when the output of $f$ has additional dimensions. Thus, WB computational complexity scales linearly with the number of added dimensions.  For example, if $q = 2d$, a single round of WB fits twice as many trees a a round fitting a GB model on the same dataset. Multi-dimension trees exist and can be used to fit WB models \citep{multitree}; however, initial tries at using multi-dimension-output trees for WB gave more time-efficient but worse overall performance.

On computation timing, it's worth noting that \texttt{wideboost} is a python wrapper around either \texttt{XGBoost} or \texttt{LightGBM}, and thus generally has more overhead per round compared to vanilla GB trained using either base package. Comparing one trial from Section 3.1 where \texttt{wideboost} is used to fit a vanilla GB model (parameters $\beta = I$, $q=d$ and $\beta$ is not estimated during training) showed that \texttt{wideboost} could take 64\% longer to train than the same model using \texttt{XGBoost} (indeed the output models were the same in this trial).  Timing can be improved with more code optimization. Currently, at least for medium-sized problems, timing is not so long that the benefits of WB shown in Section 3 become impractical.

\section{Numerical experiments}
\label{exp}

We repurpose the MNIST dataset \cite{mnist} to investigate three properties of WB:
\begin{enumerate}
    \item Can WB improve on multi-dimension output problems where the dimensions are not independent? We treat the usual multinomial outcome as separate bernoulli outcomes. WB, by not treating the multiple outcomes as independent, significantly outperforms GB which, in current implementations, treats outcomes as independent.
    \item Can WB make useful intermediate embeddings?  To look at this we use a transfer learning setup similar to the experimental setup in \cite{levin2022transfer}. An ``upstream'' model is trained to create embeddings. Quality of those embeddings is evaluated on a ``downstream'' dataset. Embeddings using WB outperform using GB predictions as an embedding (a technique better known as stacking). Embeddings from WB are shown to contain more information than GB for the modified MNIST task we consider.
    \item Does WB add value as an additional hyperparameter on usual GB problems? Paired with other hyperparameter tuning, WB was able to get to lower loss values faster than GB. This does not appear to always be the case, however. Unsurprisingly, vanilla GB is sometimes best.
\end{enumerate}

Note that all comparisons in this section compare WB to GB where both WB and GB are fit using \texttt{wideboost}. For GB models, we simply restrict the parameter space ($\beta = I$, $q=d$, $\beta$ is fixed).  This gives the most comparable results.

\subsection{Multi-dimension output learning}
\label{ex1}
We repeatedly (n = 30) fit WB and GB models on samples of the MNIST dataset, using \texttt{hyperopt} to find appropriate hyperparameter values for each WB and GB model. To shorten computation, each iteration samples new training and validation sets of size 3{,}000 and 2{,}000 respectively from the MNIST training set. The test set is a sample of size 1{,}000 from the usual MNIST test set. Features are a sample of size 200 from the usual 28$\times$28 = 784 features. \texttt{hyperopt} is given 25 chances per trial to find good hyperparameters.  \texttt{XGBoost} as a backend was given 20 early stopping rounds (based on validation performance) with 500 maximum boost rounds.

The loss calculated is average log-loss across the 10 dimension one-hot response vector. We treat the dimensions as separate binary outcomes (as opposed to the usual multinomial outcome) in order to simulate a dataset with a multivariate response and unknown dependence across the responses.

Across the 30 trials, WB makes use of the dependence across the response dimensions to achieve a statistically significant lower average log-los (see Table \ref{tab1}).
\begin{table}[htbp]
\centering
\begin{tabular}{c|r|r|r|r|r}
    Model & N Trial Wins & Avg test log-loss & t-stat & Avg n trees & t-stat\\
    \hline
    WB & 28 &  0.0541 (0.0007) & -3.9**** & 1245 (206) & -2.1*\\
    GB & 2 & 0.0582 (0.0008) &  &  1785 (145) &  \\
\end{tabular}
\caption{* t-stat is significant at the 0.05 level. **** t-stat is significant at the 0.001 level.}
\label{tab1}
\end{table}

As part of the hyperparamerter optimization, WB could have chosen plain GB (where $q = d$ and $\beta = I$). However, the hyperparameter optimization consistently chose $q > d$. For all ten trials $q > d$, and the average value for $q$ was 14.

Also of note is that WB seems to be relatively efficient with fewer trees fit on average. This is not necessarily an intuitive result because a single round of WB where $q > d$ will fit more trees than a single round in GB (for example, if $q = 14$, 40\% more trees are fit in WB compared to GB). This can be explained by WB using fewer boost rounds than GB for several trials. The median number of boost rounds used for WB was 47, while the median for GB was 153.  Roughly restating, while WB fits 140\% of the GB number of trees per round in this exercise, vanilla GB tends to use over 300\% more boost rounds.

\subsection{Embedding Quality}

Like the experiment in \ref{ex1}, we treat the MNIST response as separate binary outcomes. Following a similar experiment format to \cite{levin2022transfer}, we run several trials (n = 30) consisting of an ``upstream'' dataset used to fit an initial model on half of the binary outputs. This initial model is then used on a ``downstream'' dataset to generate embeddings of the downstream data for use in predictions on the other half of the binary outputs. The quality of the embedding is then judged by evaluating prediction performance on the downstream dataset using only the embeddings generated from the upstream model on the downstream data.

Each trial generates new upstream and downstream datasets by sampling. Upstream datasets are of size [3{,}000, 2{,}500, 2{,}000] for the train, validation and test sets respectively.  Downstream datasets are of size [500, 1{,}500, 1{,}000] for the train, valiation and test sets respectively. Each trial also samples 5 of the 10 response dimensions for the upstream dataset and leaves the remaining 5 as the response of the downstream dataset. We again use \texttt{hyperopt} to optimize the models at both the upstream and downstream stages. The hyperparamerter optimization is given 15 tries for each trial.

To ensure a fair comparison between the WB and GB embeddings, the downstream model is simply a set of GB models (one for each output dimension) using either the WB or GB embeddings as inputs.  To emphasize, even though the WB trials use WB for the upstream model, the downstream model is always a GB model.

Based on downstream test set performance, the embeddings made by WB outperform using GB logits as embeddings in a downstream model.
\begin{table}[htbp]
    \centering
    \begin{tabular}{c|r|r|r|r|r}
        Model & N Trial Wins& Avg downstream test log-loss & t-stat \\
        \hline
        WB & 30 & 0.180 (0.004) & -7.5***** \\
        GB & 0 & 0.220 (0.003) &          
    \end{tabular}
    \caption{***** t-stat is significant at less than the 0.001 level. }
\end{table}

\subsection{WB as an additional Hyperparameter}

If we employ WB head-to-head against GB on the classical MNIST prediction problem (this time treating the response properly as a multinomial), the advantages of WB are less clear.

\begin{figure}[h]
    (a)
    \begin{subfigure}[b]{0.45\textwidth}
         \centering
         \includegraphics[width=\textwidth]{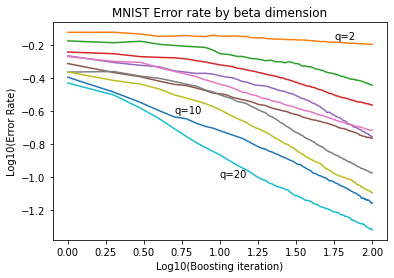}
         \label{fig:a}
    \end{subfigure}
    \hfill
    (b)
    \begin{subfigure}[b]{0.45\textwidth}
         \centering
         \includegraphics[width=\textwidth]{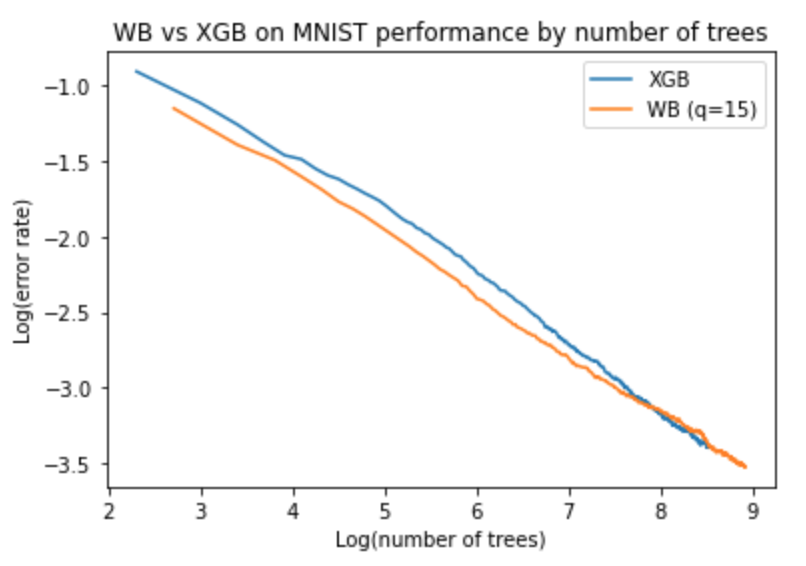}
         \label{fig:b}
     \end{subfigure}
     \label{fig:beta}
    \caption{Test error rates for differing $q$ on the the MNIST dataset. Larger $q$ shows better performance for earlier boosting rounds (subfigure (a)). However, when controlling for number of trees this difference disappears (subfigure (b)).}
    \label{fig:full}
\end{figure}

Specifying $q > d$ and leaving all other parameters the same gives better models for the same number of boosting rounds (Figure \ref{fig:full}(a)), but if you control for the extra number of trees, the benefit of WB disappears (Figure \ref{fig:full}(b)).

Nonetheless it is interesting that the two lines in Figure \ref{fig:full}(b) are so close.  On this dataset at least, the repeated gradient corrections employed in GB doesn't seem to have any special advantages over fitting more trees simultaneously and adding them together.

More generally, although on the classic task for MNIST the advantage of WB is none or negilgible, specific trials in Section 3.1 showed that WB employed with hyperparameter tuning can give better performance using less compute.

\section{Conclusion}

Wide Boosting is a relatively straightforward generalization of standard Gradient Boosting and introduces greater flexibility to the GB methodology. Specifically, WB outperformed GB on a prediction problem where the response is multi-dimensional and not statistically independent across the dimensions.  WB also introduces the concept of abstract embeddings to GB and outperformed GB on a transfer learning task. As in deep learning, these embeddings can be used for transfer learning or may be of direct interest themselves.

The python package \texttt{wideboost} makes WB publicly available. Users of the popular packages \texttt{XGBoost} and \texttt{LightGBM} in particular may find benefit in using \texttt{wideboost} because \texttt{wideboost} currently wraps both packages, supplying those packages the gradient and hessian calculations.

Given that WB can also be thought of in a computational network or Deep Learning context, there may be similar, useful network architectures that can take advantage of Gradient Boosting or other powerful prediction methodologies not traditionally related to Deep Learning. For example, future work may use gradient boosted decision trees as the first layer in deep neural networks.

\small

\bibliographystyle{apalike}
\bibliography{references.bib}

\begin{thebibliography}{}

\bibitem[Balestriero, 2017]{ndt}
Balestriero, R. (2017).
\newblock Neural decision trees.
\newblock {\em arXiv preprint arXiv:1702.07360}.

\bibitem[Bekkerman, 2015]{kddreview}
Bekkerman, R. (2015).
\newblock The present and the future of the kdd cup competition: an outsider's
  perspective.
\newblock
  \url{https://www.linkedin.com/pulse/present-future-kdd-cup-competition-outsiders-ron-bekkerman}.

\bibitem[Chen and Guestrin, 2016]{xgboost}
Chen, T. and Guestrin, C. (2016).
\newblock Xgboost: A scalable tree boosting system.
\newblock In {\em Proceedings of the 22nd acm sigkdd international conference
  on knowledge discovery and data mining}, pages 785--794.

\bibitem[Dorogush et~al., 2018]{catboost}
Dorogush, A.~V., Ershov, V., and Gulin, A. (2018).
\newblock Catboost: gradient boosting with categorical features support.
\newblock {\em arXiv preprint arXiv:1810.11363}.

\bibitem[Dumont et~al., 2009]{multitree}
Dumont, M., Marée, R., Wehenkel, L., and Geurts, P. (2009).
\newblock Fast multi-class image annotation with random subwindows and multiple
  output randomized trees.
\newblock volume~2, pages 196--203.

\bibitem[Friedman, 2001]{gbm}
Friedman, J.~H. (2001).
\newblock Greedy function approximation: a gradient boosting machine.
\newblock {\em Annals of statistics}, pages 1189--1232.

\bibitem[Grubb and Bagnell, 2011]{grubb2011generalized}
Grubb, A. and Bagnell, J.~A. (2011).
\newblock Generalized boosting algorithms for convex optimization.
\newblock In {\em Proceedings of the 28th International Conference on
  International Conference on Machine Learning}, pages 1209--1216.

\bibitem[Horrell, 2019]{blog}
Horrell, M. (2019).
\newblock Gradient fitting for deep learning.
\newblock
  \url{https://mthorrell.github.io/horrellblog/2019/04/28/gradient-fitting-for-deep-learning/}.
\newblock Accessed: 2020-07-19.

\bibitem[Ke et~al., 2017]{lightgbm}
Ke, G., Meng, Q., Finley, T., Wang, T., Chen, W., Ma, W., Ye, Q., and Liu,
  T.-Y. (2017).
\newblock Lightgbm: A highly efficient gradient boosting decision tree.
\newblock In {\em Advances in neural information processing systems}, pages
  3146--3154.

\bibitem[Ke et~al., 2019]{deepgbm}
Ke, G., Xu, Z., Zhang, J., Bian, J., and Liu, T.-Y. (2019).
\newblock Deepgbm: A deep learning framework distilled by gbdt for online
  prediction tasks.
\newblock In {\em Proceedings of the 25th ACM SIGKDD International Conference
  on Knowledge Discovery \& Data Mining}, pages 384--394.

\bibitem[Kontschieder et~al., 2015]{dndf}
Kontschieder, P., Fiterau, M., Criminisi, A., and Rota~Bulo, S. (2015).
\newblock Deep neural decision forests.
\newblock In {\em Proceedings of the IEEE international conference on computer
  vision}, pages 1467--1475.

\bibitem[LeCun and Cortes, 2010]{mnist}
LeCun, Y. and Cortes, C. (2010).
\newblock {MNIST} handwritten digit database.
\newblock \url{http://yann.lecun.com/exdb/mnist/}.

\bibitem[Levin et~al., 2022]{levin2022transfer}
Levin, R., Cherepanova, V., Schwarzschild, A., Bansal, A., Bruss, C.~B.,
  Goldstein, T., Wilson, A.~G., and Goldblum, M. (2022).
\newblock Transfer learning with deep tabular models.
\newblock {\em arXiv preprint arXiv:2206.15306}.

\bibitem[Microsoft, 2020]{microsoft}
Microsoft (2020).
\newblock Lightgbm: Machine learning challenge winning solutions.
\newblock
  \url{https://github.com/microsoft/LightGBM/tree/master/examples#machine-learning-challenge-winning-solutions}.

\bibitem[Popov et~al., 2019]{node}
Popov, S., Morozov, S., and Babenko, A. (2019).
\newblock Neural oblivious decision ensembles for deep learning on tabular
  data.
\newblock {\em arXiv preprint arXiv:1909.06312}.

\end{thebibliography}

\section{Appendix: Gradient and Hessian calculations}
\label{gradhes}

Consider the $i$-th row of $f(X)$, denoted $f_{[i]}$. We calculate $\frac{\partial L}{\partial f_{[i]}}$ and $\frac{\partial L}{\partial f_{[i]} \partial f_{[i]}^T}$, where $L$ is defined in (\ref{wb}). If we denote $G_{[i]}$ and $H_{[[i]]}$ as the gradient and hessian of $L$ with respect to the $i$-th row of $f(X)$ when $L$ is defined by (\ref{loss}), the gradient and hessians for $L$ with respect to $f_{[i]}$ when $L$ is defined by (\ref{wb}) can be computed using the chain rule:

\begin{eqnarray}
\frac{\partial L}{\partial f_{[i]}} &=& G_{[i]} \beta^T \nonumber \\
\frac{\partial L}{\partial f_{[i]} \partial f_{[i]}^T} &=& \beta H_{[[i]]} \beta^T \nonumber
\end{eqnarray}
Applying these formulas to common loss functions we find the following example  calculations for gradients and hessians for Wide Boosting. Note we will use the subscript $_{[i]}$ to denote the $i$-th row of a matrix. 

\subsubsection{Regression -- Squared Error}
$Y \in \mathbb{R}^{n \times 1}$, $f(X) \in \mathbb{R}^{n \times q}$ and $\beta \in \mathbb{R}^{q \times 1}$. 

\begin{eqnarray}
L(Y,f(X)\beta) & = & \frac{1}{2} \| Y - f(X)\beta \|^2  \nonumber \\
\frac{\partial L}{\partial f_{[i]} } & = & (f(X)\beta - Y)_{[i]} \beta^T \nonumber \\
\frac{\partial^2 L}{\partial f_{[i]} \partial f_{[i]}^T} &=& \beta \beta^T \nonumber
\end{eqnarray}

\subsubsection{Binary Classification -- Log-loss}
$Y \in \{0,1\}^{n \times 1}$, $f(X) \in \mathbb{R}^{n \times q}$ and $\beta \in \mathbb{R}^{q \times 1}$

\begin{eqnarray}
\mbox{Let} ~ P \in \mathbb{R}^{n \times 1} ~\mbox{where}~ P_{[i]} &=& \frac{\exp[f(X)\beta]_{[i]}}{1 + \exp[f(X)\beta]_{[i]}}  \nonumber \\
L(Y,f(X)\beta) & = & Y^T \log(P) + (\mathbf{1}-Y)^T \log(\mathbf{1}-P) \nonumber \\
\frac{\partial L}{\partial f_{[i]} } & = & (P - Y)_{[i]} \beta^T \nonumber \\
\frac{\partial^2 L}{\partial f_{[i]} \partial f_{[i]}^T} &=& \left(P_{[i]} - P_{[i]}^2\right) \beta \beta^T \nonumber
\end{eqnarray}
where the $\exp[ \cdot ]$ and $\log[\cdot]$ functions are applied elementwise to the input matrix and $\mathbf{1} \in \mathbb{R}^{n\times1}$ is a vector of all ones. 

\subsubsection{Multi-class Classification -- Log-loss}
$Y \in \{0,1\}^{n \times d}$, $f(X) \in \mathbb{R}^{n \times q}$ and $\beta \in \mathbb{R}^{q \times d}$

\begin{eqnarray}
\mbox{Let} ~ P \in \mathbb{R}^{n \times d} ~\mbox{where}~ P_{[i]} &=& \frac{\exp[f(X) \beta]_{[i]}}{\exp [f(X)\beta]_{[i]}  \mathbf{1}} \nonumber \\
L(Y,f(X)\beta) &=& \mbox{tr}(Y \log(P)^T)  \nonumber \\
\frac{\partial L}{\partial f_{[i]} }  & = & (P - Y)_{[i]} \beta^T \nonumber \\
\frac{\partial^2 L}{\partial f_{[i]} \partial f_{[i]}^T}  &=&
\beta \left( \mbox{diag}[P_{[i]}] - P_{[i]}^T P_{[i]} \right) \beta^T
\nonumber
\end{eqnarray}
where the $\exp[ \cdot ]$ and $\log[\cdot]$ functions are applied elementwise to the input matrix and $\mathbf{1} \in \mathbb{R}^{d\times1}$ is a vector of all ones.  The $\mbox{diag}[\cdot]$ function takes a row vector and returns a square, diagonal matrix with diagonal elements corresponding to the elements of the vector.

\end{document}